\pdfoutput=1
\documentclass[11pt]{article}

\usepackage[]{ACL2023}

\usepackage{times}
\usepackage{latexsym}

\usepackage[T1]{fontenc}
\usepackage[utf8]{inputenc}

\usepackage{microtype}

\usepackage{inconsolata}
\usepackage{booktabs}
\usepackage{makecell}
\usepackage{amsfonts}
\usepackage{amsmath}

\usepackage{listings}
\usepackage{graphicx}
\usepackage{todonotes}
\usepackage{multirow}

\lstdefinestyle{mystyle}{
    basicstyle=\ttfamily\footnotesize,
    breakatwhitespace=false,         
    breaklines=true,                 
    breakatwhitespace=true,
    captionpos=b,                    
    keepspaces=true,                 
    numbersep=5pt,                  
    showspaces=false,                
    showstringspaces=false,
    showtabs=false,                  
    tabsize=2,
    postbreak=\mbox{\textcolor{red}{$\hookrightarrow$}\space},
}

\newcommand{\strict}{\textsc{strict}}
\newcommand{\strictsmall}{\textsc{strict-small}}
\newcommand{\ourmodel}{BLaLM}

\title{Sample-Efficient Language Modeling with Linear Attention and Lightweight Enhancements}

\author{Patrick Haller \And Jonas Golde \\
\\
Humboldt-Universität zu Berlin
\\
\texttt{\{patrick.haller.1{\normalfont,} jonas.max.golde{\normalfont,} alan.akbik\}@hu-berlin.de} \And
Alan Akbik \\
}

\begin{document}
\maketitle
% \begin{abstract}
% We study architectural and optimization techniques for sample-efficient language modeling under the constraints of the BabyLM 2025 shared task. Our model, \ourmodel{}, replaces self-attention with a linear-time mLSTM token mixer. We evaluate lightweight enhancements, including short convolutions, sliding window attention with dynamic modulation, and Hedgehog feature maps. To support training in low-resource settings, we curate a high-quality corpus emphasizing readability, coherence, and pedagogical structure. Experiments on both strict and strict-small tracks show that \ourmodel{} outperforms Transformer baselines in the low-data regime and remains competitive at larger scales. We further compare the optimizers AdamW and Muon, finding that Muon improves convergence and training stability. Our findings demonstrate that targeted modifications can yield strong performance without additional data or supervision.
% \end{abstract}

\begin{abstract}
We study architectural and optimization techniques for sample-efficient language modeling under the constraints of the BabyLM 2025 shared task. Our model, \textbf{BLaLM}, replaces self-attention with a linear-time mLSTM token mixer and explores lightweight enhancements, including short convolutions, sliding window attention with dynamic modulation, and Hedgehog feature maps. To support training in low-resource settings, we curate a high-quality corpus emphasizing readability and pedagogical structure. Experiments across both \strict{} and \strictsmall{} tracks show that (1) linear attention combined with sliding window attention consistently improves zero-shot performance, and (2) the Muon optimizer stabilizes convergence and reduces perplexity over AdamW. These results highlight effective strategies for efficient language modeling without relying on scale.
\end{abstract}

\section{Introduction}

Training language models under strict resource constraints remains a central challenge, both for advancing theoretical understanding and for enabling practical deployment on limited hardware. The BabyLM shared task provides a unique opportunity to evaluate models in a controlled setting, where participants are restricted to training on at most 10 million (\strictsmall) or 100 million (\strict) words for a maximum of 10 epochs. This environment encourages the development of sample-efficient algorithms rather than scale-dependent strategies.

Our submission focuses on algorithmic enhancements rather than introducing novel architectures. Specifically, we examine whether recent advancements in model design and optimization can be adapted to improve sample efficiency when applied to a standard Transformer backbone. Our contributions are as follows:

\begin{enumerate}
    \item \textbf{Model Architecture:} We replace the self-attention mechanism in a standard Transformer with the linear-time mLSTM module, yielding an efficient subquadratic variant we refer to as \ourmodel{}.
    \item \textbf{Optimization:} We evaluate the \textit{Muon} optimizer, a recently proposed alternative to AdamW, which introduces dynamic momentum and a decoupled weight decay schedule. We compare Muon and AdamW under identical training conditions.
    \item \textbf{Architectural Enhancements:} We introduce and evaluate several lightweight modifications to the \ourmodel{} model, including sliding window attention (SWA), short convolutional layers, and dynamic attention modulation.
    \item \textbf{Corpus Construction:} We curate a high-quality corpus by filtering and modifying existing text corpora, aiming to improve training dynamics for small models. Preliminary results indicate improved downstream performance relative to unfiltered datasets.
\end{enumerate}

Our experiments lead to two key findings: First, replacing self-attention with a linear-time mLSTM token mixer, especially when combined with sliding window attention and dynamic modulation, leads to strong zero-shot performance under low-resource constraints. Second, the Muon optimizer improves convergence and stability compared to AdamW, particularly for matrix-shaped parameters. Together, these results point to practical strategies for improving sample efficiency in compact language models.

\section{Preliminaries and Related Work}

\subsection*{Transformers}
The Transformer architecture, proposed by~\citet{vaswani2017}, has become the de facto standard for large-scale language modeling. Unlike recurrent neural networks (RNNs) or long short-term memory networks (LSTMs)~\citep{lstm}, Transformers process sequential input in parallel through self-attention. Given query, key, and value matrices $\mathbf{Q}, \mathbf{K}, \mathbf{V} \in \mathbb{R}^{n \times d}$, the self-attention output is computed as:
\begin{equation}
    \boldsymbol{y} = \text{softmax}\left(\frac{\boldsymbol{Q} \boldsymbol{K}^\top}{\sqrt{d_k}} \odot \boldsymbol{M}\right) \boldsymbol{V},
\end{equation}
where $\mathbf{M}$ is a causal mask that prevents attending to future tokens. While highly expressive, self-attention incurs $\mathcal{O}(n^2 d)$ complexity in both computation and memory, which becomes a bottleneck for long sequences, especially during autoregressive decoding.

\subsection*{Linear Attention}
To address the quadratic bottleneck, \citet{katharopoulos2020transformersrnnsfastautoregressive} proposed linear attention mechanisms that replace the softmax kernel with a feature map $\phi(\cdot)$ such that:
\begin{equation}
    \text{softmax}(\mathbf{Q}\mathbf{K}^\top) \approx \phi(\mathbf{Q}) \phi(\mathbf{K})^\top.
\end{equation}
This formulation enables autoregressive decoding in $\mathcal{O}(n d^2)$ time by exploiting the associativity of matrix multiplication, reducing memory usage and improving scalability.

Linear attention has been extended in numerous architectures aimed at long-context modeling and efficient training~\citep{sun2023retentivenetworksuccessortransformer, poli2023hyenahierarchylargerconvolutional}. Owing to their subquadratic inference efficiency, several works~\citep{bick2025transformersssmsdistillingquadratic, lan2025ligerlinearizinglargelanguage, haller2025empiricalevaluationknowledgedistillation, vannguyen2025lizardefficientlinearizationframework} focus on linearizing Transformer-based language models into subquadratic architectures.
In the BabyLM 2024 shared task, \citet{haller-etal-2024-babyhgrn} introduced \textit{BabyHGRN}, which leverages a recurrent HGRN2 token mixer within Transformer-style blocks. It achieved competitive results under low-resource constraints, motivating continued exploration of subquadratic alternatives.

\subsection*{xLSTM and mLSTM}
xLSTM~\citep{beck2024xlstmextendedlongshortterm} revisits the LSTM architecture with two core innovations: exponential gating and enhanced memory structures. It defines two cells, sLSTM and mLSTM, which are assembled into residual blocks.

\textbf{mLSTM} extends the scalar memory $c_t$ to a matrix memory $\mathbf{C}_t \in \mathbb{R}^{d \times d}$ that stores key-value pairs via an outer-product update. The forget gate $f_t$ acts as a decay, while the input gate $i_t$ controls the learning rate:
\begin{equation}
    \mathbf{C}_t = f_t \mathbf{C}_{t-1} + i_t v_t k_t^\top, \quad n_t = f_t n_{t-1} + i_t k_t,
\end{equation}
and retrieval is computed using:
\begin{equation}
    h_t = o_t \odot \frac{\mathbf{C}_t q_t}{\max\{|\langle n_t, q_t \rangle|, 1\}},
\end{equation}
with $q_t, k_t, v_t$ derived from learned projections.

As with other linear-time mechanisms, mLSTM supports parallel training and linear-time autoregressive decoding. It serves as the token mixer in our model architecture.

\subsection*{The BabyLM Benchmark}
The BabyLM initiative~\citep{charpentier2025babylmturns3papers} introduced a suite of benchmarks for evaluating language models in low-resource conditions, with a focus on learnability, generalization, and alignment with developmental stages. The 2025 shared task continues this focus, imposing strict limits on training data and epochs to emphasize sample efficiency and high quality data curation.

\subsection*{Optimizers}
Adaptive optimizers such as Adam~\citep{kingma2017adammethodstochasticoptimization} and its decoupled variant AdamW~\citep{loshchilov2019decoupledweightdecayregularization} remain standard for LLM training due to their robustness and ease of tuning. However, their dynamics can be suboptimal for matrix-shaped parameters, especially in low-data or large-batch regimes.

Recent alternatives aim to improve convergence and stability, including Lion~\citep{chen2023symbolicdiscoveryoptimizationalgorithms}, Sophia~\citep{liu2024sophiascalablestochasticsecondorder}, and Shampoo~\citep{gupta2018shampoopreconditionedstochastictensor}. Muon~\citep{keller2024muon} orthogonalizes gradient updates via a truncated Newton-Schulz iteration, improving conditioning for matrix-valued parameters with minimal overhead. It is typically used in hybrid schemes, where scalar parameters (e.g., layer norms, biases) are still optimized with AdamW.

Muon has shown benefits in both vision and language domains~\citep{ai2025practicalefficiencymuonpretraining, liu2025muonscalablellmtraining}, including better training stability, faster convergence, and improved data efficiency, which all are valuable under the constraints of BabyLM.

\begin{table*}[ht]
    \centering
    \begin{tabular}{lrr}
    \toprule
    Dataset & \# Words \strictsmall & \# Words \strict \\
    \midrule
    CHILDES Project (Child-directed speech) & 2M & 8.7M \\
    Fineweb-Edu  & 2M & 21M \\
    TinyStories & 1M & 35M \\
    Project Gutenberg, Fiction Books & 1.5M & 1.7M \\
    Simple Wikipedia (English) & 1.5M & 22.6M \\
    Cosmopedia & & \\
    $\text{\ \ \ \ - WikiHow}$  & 1.8M & 10.1M \\
    $\text{\ \ \ \ - Math}$ & 0.2M & 0.3M \\
    
    \midrule
    \textit{Total} & $\approx $ 10M & $\approx$ 99.5M\\
     \bottomrule
    \end{tabular}
    \caption{Token counts per data source in the curated corpus used for the \strictsmall{} and \strict{} tracks of BabyLM 2025.}
    \label{tab:dataset_overview}
\end{table*}

\section{Data Curation}\label{sec:dataset}
Data quality plays a critical role in small-scale language modeling, where noisy or incoherent samples can substantially degrade performance. In this work, we prioritize readability, coherence, and syntactic simplicity to improve learnability under low-resource constraints.

Rather than relying solely on large, unfiltered corpora, we curate a dataset by filtering and modifying existing sources using heuristic and LLM-guided approaches. Our filtering pipeline targets syntactically clean, semantically rich, and pedagogically structured documents likely to be learnable by small models.
\subsection{Data Sources}

Our curated pretraining corpus draws from a diverse set of publicly available datasets selected for their relevance to early language acquisition, general knowledge, and structured instruction. The largest component is \textbf{FineWeb-Edu}~\citep{lozhkov2024finewebedu, penedo2025fineweb2pipelinescale}, a filtered subset of FineWeb-2 annotated for educational value. To incorporate spoken language patterns, we include transcripts from the \textbf{CHILDES} corpus~\citep{macwhinney2000childes}, which features child-directed speech. We also leverage \textbf{TinyStories}~\citep{eldan2023tinystoriessmalllanguagemodels}, a synthetic story dataset designed for early learners. Fictional content is sourced from a filtered selection of English novels from \textbf{Project Gutenberg}~\citep{e22010126}, while simplified encyclopedic entries come from \textbf{Simple Wikipedia}. Finally, we include domain-specific educational content from \textbf{Cosmopedia}~\citep{benallal2024cosmopedia}, which covers instructional materials such as WikiHow articles and mathematics explanations. A breakdown of word counts per dataset and track is provided in Table~\ref{tab:dataset_overview}.

\begin{figure*}[ht!]
    \centering
    \includegraphics[width=0.9\linewidth]{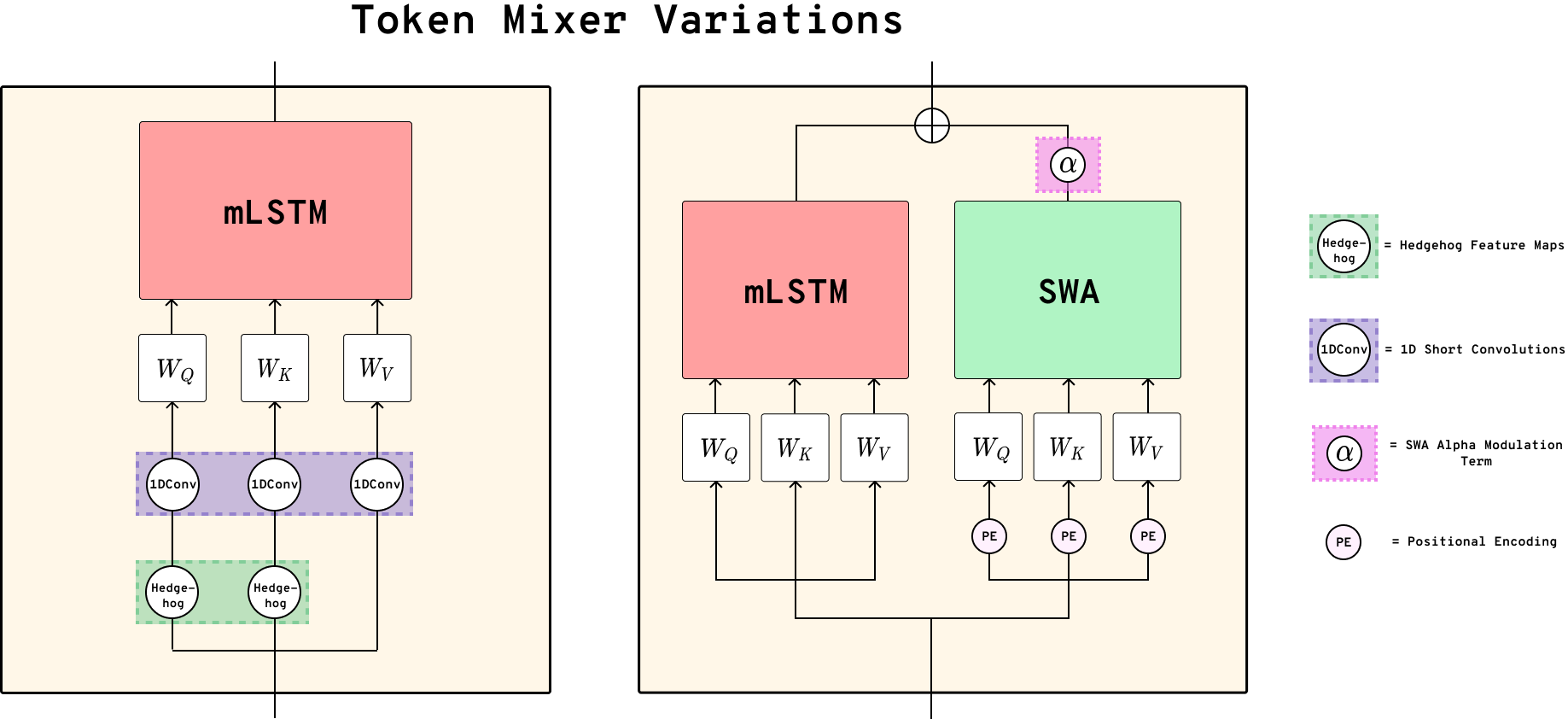}
    \caption{Overview of the \ourmodel{} architecture. The standard self-attention module is replaced by an mLSTM token mixer. Optional enhancements such as sliding window attention (SWA) can be integrated and combined with mLSTM outputs.}
    \label{fig:overview}
\end{figure*}
\subsection{Filtering Pipeline}

We apply dataset-specific filters to improve linguistic quality and reduce noise. Below we summarize our main filtering strategies:

\paragraph{FineWeb-Edu}
Although FineWeb-Edu is already annotated for educational value, we re-evaluate all samples using our own educational scoring prompt (Appendix~\ref{appendix:prompt}) with LLaMA 3.3–70B. % This allows us to define a consistent scoring rubric across all sources and to apply stricter filtering criteria based on both educational content and readability.

\paragraph{Gutenberg Fiction}
We discard Gutenberg entries without named entities and subsample up to 200 samples per book to ensure diversity.

\paragraph{TinyStories}
We remove template-like introductions (e.g., “Once upon a time…”) to reduce repetition and increase stylistic variety.

\paragraph{Simple Wikipedia}
We retain only paragraphs with at least 15 words to remove boilerplate and fragmented content.

\paragraph{Cosmopedia (WikiHow \& Math)}
We filter for content relevant to K–12 learners and remove excessively long or domain-specific passages.

\paragraph{CHILDES (Child-Directed Speech)}
CHILDES contains transcripts of parent–child dialogue, marked with speaker tags (e.g., “*MOT:” for mother). We apply:
\begin{enumerate}
    \item \textbf{Speaker Tag Removal:} Prefixes like “*MOT:” or “*COL:” are removed.
    \item \textbf{Minimum Length Filtering:} We discard utterances with fewer than 7 words.
    \item \textbf{Grammar Correction:} We normalize speech using LanguageTool for improved grammaticality.
\end{enumerate}

All data is tokenized and counted at the word level to ensure the final corpus respects the BabyLM 2025 limits of 10M (\strictsmall{}) and 100M (\strict{}) words.

\section{Model Architecture and Optimization}

We aim to evaluate whether architectural and optimization strategies known to improve large-scale language models can also improve sample efficiency under strict training budgets. Rather than designing a novel architecture, we incrementally modify a standard Transformer decoder to assess the contribution of individual components.

Our model, referred to as \textbf{\ourmodel{}} (\textbf{B}aby \textbf{L}inear \textbf{A}ttention \textbf{LM}), follows the general architecture of recent Qwen models~\citep{bai2023qwentechnicalreport}. It uses \textbf{pre-normalization} with \textbf{RMSNorm}~\citep{zhang2019rootmeansquarelayer} for training stability, \textbf{feed-forward blocks} with SwiGLU activations, and \textbf{rotary positional embeddings (RoPE)}~\citep{su2023roformerenhancedtransformerrotary} to encode position information. RoPE is used in both the self-attention baseline and the optional sliding window attention modules in \ourmodel{}.

The key deviation from the standard Transformer lies in the \textbf{token mixer}, which is the module responsible for integrating contextual information across tokens. In Transformers, this role is fulfilled by the self-attention mechanism; in \ourmodel{}, we replace it with \textbf{mLSTM}, a recurrent linear-time alternative. The mLSTM operates via element-wise gating (forget and input gates) and uses matrix-valued memory updates across learned projections. It supports fully parallel training and linear-time autoregressive decoding, thereby avoiding the quadratic overhead of softmax attention while maintaining expressivity.

This architectural choice preserves full compatibility with Transformer training pipelines, allowing direct comparisons between self-attention and mLSTM-based token mixing.

\begin{table*}[ht]
    \centering
    \begin{tabular}{lccccccc}
    \toprule
    \textsc{Dataset} & \textsc{BLiMP} & \textsc{B. Suppl.} & \textsc{Entity} & \textsc{EWoK} & \textsc{Eye} & \textsc{Reading} & \textsc{Avg.} \\
    & acc. & acc. & acc. & acc. & $\Delta R^2$ & $\Delta R^2$ & \\
    \midrule
    Baseline-10M       & 64.96 & 66.8 & 40.01 & 51.55 & 0.98 & 0.45 & 37.45 \\
    Baseline-100M      & 75.68 & 65.2 & 34.82 & 51.82 & 0.82 & 0.33 & \textbf{38.11} \\
    Our-10M            & 67.99 & 63.6 & 39.46 & 52.27 & 1.09 & 0.65 & 37.51 \\
    Our-100M           & 74.51 & 57.6 & 14.65 & 55.73 & 1.18 & 0.90 & 34.09 \\
    \bottomrule
    \end{tabular}
    \caption{Comparison of the official BabyLM dataset ("Baseline") and our curated corpus ("Ours") across both strict-small and strict tracks. We report average zero-shot performance; full results in Appendix~ \ref{appendix:dataset_comparison_full_results}.}
    \label{tab:dataset_comparison}
\end{table*}

\subsection{Architectural Enhancements}

In addition to the mLSTM substitution, we introduce a set of lightweight architectural improvements aimed at enhancing sample efficiency:
\begin{itemize}
    \item \textbf{Short Convolutions (ShortConv):} 1D depthwise convolutions are added before the token mixer on the query and key projections to enhance local inductive bias. Recently added by \citet{gu2024mambalineartimesequencemodeling,dao2024transformersssmsgeneralizedmodels,beck2024xlstmextendedlongshortterm,lan2025ligerlinearizinglargelanguage,vannguyen2025lizardefficientlinearizationframework}.
    \item \textbf{Sliding Window Attention (SWA)}~\citep{beltagy2020longformerlongdocumenttransformer}: A local attention mechanism with fixed-size attention window. Sliding is used in conjunction with the mLSTM token mixer. The input is passed through both modules and added together, like:
    \begin{equation}
        h_{final} = \frac{h_{LA}}{2} + \frac{h_{SWA}}{2}
    \end{equation}
    \item \textbf{SWA with Dynamic Modulation (DynMod):} Applies a learned gating function to modulate attention hidden states over each layer.
    \begin{equation}
        h_{total} = h_{LA} + \alpha \cdot h_{SWA}
    \end{equation}
    \begin{equation}
        h_{total} = h_{LA} + tanh(\alpha) \cdot h_{SWA}
    \end{equation}
    \item \textbf{Hedgehog Feature Maps}~\citep{zhang2024hedgehogporcupineexpressive}: A recently proposed mechanism  that mimics several properties of softmax-based attention. It is applied to the query and key projections.
\end{itemize}

Each mechanism, as illustrated in Figure~\ref{fig:overview}, is introduced independently and evaluated against the base \ourmodel{}~configuration to quantify its contribution under fixed training budgets.

\section{Training Setup}
\label{sec:training_setup}

All experiments are conducted under the BabyLM 2025 shared task constraints for the \strictsmall{} (10M words) and \strict{} (100M words) tracks, using the curated dataset described in Section~\ref{sec:dataset}

\subsection*{Sequence Length and Batching}
We train with a context length of 512 tokens and an effective global batch size of 64. When hardware limitations require smaller per-device batches, we use gradient accumulation to match the target batch size. Text data is first concatenated into a continuous stream before splitting into fixed-length sequences to avoid truncation and minimize padding overhead.

\subsection*{Models}
We use two architectural variants throughout our experiments: a baseline Transformer decoder (Qwen-style) and our proposed model, \ourmodel{}, which replaces self-attention with an mLSTM token mixer. Both models share the same configuration where applicable; architectural differences are detailed in Appendix~\ref{appendix:model_configuration}.

\subsection*{Training Duration and Checkpointing}
Each model is trained for a maximum of 10 epochs over the respective corpus. We evaluate all saved checkpoints and report results for the best-performing one based on average downstream performance.

\subsection*{Evaluation}
All models, except the final submissions, are evaluated using the fast zero-shot evaluation suite provided by the BabyLM organizers~\citep{charpentier2025babylmturns3papers}.\footnote{Shortly before the deadline, a bug was discovered in the evaluation for the \textit{WuG} task. Due to time constraints, we were unable to re-evaluate all models. We therefore exclude this task from our reported results.}  
We rely on this fast evaluation method to score all intermediate checkpoints and select the best model per run. Final submissions are evaluated on hidden tasks and additionally fine-tuned on GLUE.\footnote{For a complete list of benchmarks and descriptions, see the \href{https://github.com/babylm/evaluation-pipeline-2025}{official BabyLM 2025 evaluation pipeline}.}

\subsection*{Learning Rate Scheduling}
We use a cosine decay schedule with a 10\% linear warmup phase. The learning rate used for each experiment is reported in the corresponding results section.

\subsection*{Optimizers}
\label{sec:optimizer}
We use either AdamW or Muon, as introduced in Section~2. In Sections~\ref{exp:dataset} and~\ref{exp:architecture}, AdamW is used as the default optimizer. Later experiments switch to Muon, which is applied to matrix-shaped parameters (e.g., projection weights, MLP layers), while AdamW handles all scalar-valued parameters (e.g., embeddings, biases, and normalization layers).

\begin{table*}[ht]
    \centering
    \resizebox{\textwidth}{!}{
    \begin{tabular}{cll|ccccccc}
    \toprule
    \textsc{Track} & \textsc{Model} & \textsc{LR} & \textsc{BLiMP} & \textsc{B. Suppl.} & \textsc{Entity} & \textsc{EWoK} & \textsc{Eye} & \textsc{Reading} & \textsc{Avg.} \\
    & & & acc. & acc. & acc. & acc. & $\Delta R^2$ & $\Delta R^2$ & \\
    \midrule
    \multirow{2}{*}{Strict-Small}
        & Transformer (9) & 4e-4 & 64.95 & 57.2 & 18.07 & 51.36 & 1.13 & 0.93 & 32.27 \\
        & \ourmodel{} (9)   & 5e-4 & 66.72 & 55.6 & 40.93 & 51.0 & 0.91 & 0.60 & 35.96 \\
    \midrule
    \multirow{2}{*}{Strict}
        & Transformer (9) & 4e-4 & 72.44 & 62.0 & 20.63 & 53.36 & 1.02 & 0.74 & 35.03 \\
        & \ourmodel{} (10)  & 5e-4 & 74.49 & 60.4 & 21.99 & 53.91 & 1.03 & 0.71 & 35.42 \\
    \bottomrule
    \end{tabular}
    }
    \caption{Zero-shot performance comparison between Transformer and \ourmodel{} across both BabyLM tracks. Results reflect the best-performing epoch per model in brackets after the model name. See Appendix D for full details.}
    \label{tab:architecture_comparison}
\end{table*}
% \begin{table}[]
%     \centering
%     \begin{tabular}{lcc}
%     \toprule
%         Architecture & \textsc{Strict-Small} & \textsc{Strict} \\
%     \midrule
%         Transformer & 32.27 & 35.03 \\
%         mLSTM & 35.96 & 35.42  \\
%         \bottomrule
%     \end{tabular}
%     \caption{Caption}
%     \label{tab:my_label}
% \end{table}

\section{Experiments}
\subsection{Experiment 1: Dataset Performance}
\label{exp:dataset}

This experiment evaluates the impact of our curated dataset relative to the baseline corpus provided by the BabyLM organizers. Since the original training configuration of the baseline models could not be fully replicated, particularly in terms of preprocessing, we train our own baseline models using their corpus under our experimental setup for a fair comparison.

\paragraph{Setup}
We use our proposed architecture (\ourmodel{}) and train two variants on each dataset, the BabyLM-provided corpus and our curated corpus, for both the \strictsmall{} and \strict{} tracks. Each configuration is run twice with identical hyperparameters to control for variance. To keep the comparison controlled, we fix the learning rate at $4 \times 10^{-4}$ for all runs.

\paragraph{Results}
Table~\ref{tab:dataset_comparison} shows that in the \strictsmall{} setting, our dataset yields slightly higher average scores (37.51 vs. 37.45), with improvements observed in \textsc{BLiMP}, \textsc{EWoK}, and \textsc{Entity} accuracy. In the \strict{} track, the performance gap reverses, the baseline corpus outperforms ours, particularly on \textsc{BLiMP Supplement} and \textsc{Entity}.

These results suggest that dataset quality plays a stronger role in low-resource settings, where clean, coherent input provides better learning signals for small models. While the curated data does not consistently outperform the baseline at larger scales, it performs on par, and slightly better in the strict-small regime, without requiring additional sources or augmentation.

Because this dataset was specifically optimized for educational quality, readability, and structure, we use it for all subsequent experiments.

\subsection{Experiment 2: Transformers vs. Linear Attention}
\label{exp:architecture}

This experiment assesses the effect of replacing the standard self-attention mechanism in a Transformer with an mLSTM-based token mixer.

\paragraph{Setup}
We compare two architectures: a baseline Transformer decoder (following the Qwen configuration) and our proposed model, \ourmodel{}. Both models share the same configuration where applicable, differing only in the token mixer. Due to small differences in parameterization between self-attention and mLSTM, the number of layers is adjusted to keep parameter counts approximately matched. Full architectural details are provided in Appendix~\ref{appendix:model_configuration}.

Experiments are conducted for both the \strictsmall{} and \strict{} tracks. For each architecture, we train models using three learning rates (3e-4, 4e-4, 5e-4) to account for differences in convergence dynamics.

\paragraph{Results}
Table~\ref{tab:architecture_comparison} presents the evaluation results. In the \strictsmall{} setting, \ourmodel{} consistently outperforms the Transformer baseline across all learning rates, with the best configuration (5e-4) improving the average score from 32.27 to 35.96.

In the \strict{} track, results are more balanced. While the Transformer baseline performs better at some learning rates, \ourmodel{} achieves the highest overall score (35.42 compared to 35.03) showing that the benefits of linear attention persist even in the presence of more data, albeit with smaller margins.

These results support the hypothesis that linear-time alternatives like mLSTM can improve sample efficiency in the low-data regime and remain competitive at larger scales, making them a viable drop-in replacement for self-attention in resource-constrained training scenarios.

\subsection{Experiment 3: The Choice of Optimizer}
\label{exp:optimizer}

This experiment compares two optimizers for pretraining \ourmodel{}: AdamW, the default choice for Transformer training, and Muon, a recently proposed optimizer designed to improve convergence speed and numerical conditioning for matrix-valued parameters.

\paragraph{Setup}
AdamW is applied to all parameters, while Muon is used in a hybrid scheme as described in Section~\ref{sec:optimizer}. Specifically, Muon updates matrix-shaped parameters such as projections and MLP weights, while scalar-valued parameters (e.g., biases, embeddings, normalization layers) are handled by AdamW. 

Experiments are conducted in the \strict{} track using a fixed learning rate of 4e-4. Each optimizer is evaluated across three independent runs to account for variability in training and initialization. Performance is measured both in terms of validation perplexity and average zero-shot score.

\begin{table}[h]
    \centering
    \begin{tabular}{lcc}
    \toprule
    \textsc{Optimizer} & \textsc{PPL} & \textsc{Avg.} \\
    \midrule
    AdamW & 11.21 $\pm 0.11$ & 35.75 $\pm 1.74$ \\
    Muon & 7.95 $\pm0.15$ & 36.24 $\pm 1.16$ \\
    \bottomrule
    \end{tabular}
    \caption{Validation perplexity and average zero-shot scores across three runs comparing AdamW and Muon optimizers for xLSTM training.}
    \label{tab:optimizer_results_overall}
\end{table}

\paragraph{Results}
Table~\ref{tab:optimizer_results_overall} summarizes the results. Muon achieves a lower average validation perplexity (7.95 ± 0.15) compared to AdamW (11.21 ± 0.11), suggesting more stable and efficient optimization.

Zero-shot performance is slightly higher for Muon (36.24 ± 1.16) than for AdamW (35.75 ± 1.74), although the gap is modest. Notably, Muon exhibits more consistent results across runs, indicating improved training stability.

Overall, these findings suggest that Muon improves convergence and may lead to marginal downstream gains under strict resource constraints. Based on these observations, we use Muon for all subsequent experiments.

\subsection{Experiment 4: Learning Rate Sweep}

This experiment aims to identify the optimal learning rate for pretraining \ourmodel{} under the BabyLM constraints for both the \strictsmall{} and \strict{} tracks.

\paragraph{Setup}
We conduct a sweep over learning rates in the range from 2e-4 to 7e-4. Each configuration is trained using the same setup described in Section~\ref{sec:training_setup}, with Muon as the optimizer and a training budget of 10 epochs.

After the initial sweep, we include one additional intermediate learning rate for each track, selected based on observed trends in the initial results. All models are evaluated based on validation perplexity and average zero-shot score. Full results are provided in Appendix~\ref{appendix:lr_sweep}.

\begin{table}
    \centering
    \resizebox{0.45\textwidth}{!}{
    \begin{tabular}{lcccc}
    \toprule
    & \multicolumn{2}{c}{\strictsmall} & \multicolumn{2}{c}{\strict} \\
    \textsc{Learning Rate} & \textsc{PPL.} & \textsc{Avg.} & \textsc{PPL.} & \textsc{Avg.} \\
    \midrule
    2e-4 & 16.41 & 34.80 & 9.83 & 34.28 \\
    3e-4 & 16.41 & 35.61 & 8.46 & 35.82 \\
    4e-4 & 20.01 & 37.27 & 8.06 & 35.08 \\
    5e-4 & 16.41 & 35.03 & 7.74 & 35.82 \\
    6e-4 & 16.61 & 34.17 & 7.76 & 35.06 \\
    7e-4 & \textbf{15.73} & \textbf{37.53} & 7.64 & 36.10 \\
    \midrule
    \textit{Additional Learning Rates} \\
    7.5e-4    & 14.84 & 37.04 & - & - \\
    5.5e-4 & - & - & 7.70 & \textbf{37.49} \\
    \midrule
    \bottomrule
    \end{tabular}}
    \caption{Results from a learning rate sweep for \ourmodel{} on both tracks. Additional intermediate rates were selected based on observed trends.}
    \label{tab:lr_sweep}
\end{table}
\paragraph{Results}
Table~\ref{tab:lr_sweep} reports evaluation results for all tested learning rates. In the \strictsmall{} track, the highest average score is achieved at 7e-4 (37.53), while 4e-4 and 5e-4 also perform competitively. A follow-up experiment with 7.5e-4 yields slightly lower performance (37.04), suggesting diminishing returns beyond 7e-4.

In the \strict{} track, performance peaks at 5.5e-4 with an average score of 37.49.
This outperforms 5e-4 and 7e-4, suggesting 5.5e-4 offers the best trade-off. 

Overall, the results highlight that optimal learning rates differ by data scale. In low-resource regimes, higher learning rates such as 7e-4 are beneficial, while in higher-resource settings, more moderate values around 5.5e-4 provide the best trade-off between stability and generalization.

\subsection{Experiment 5: Evaluating Lightweight Architectural Enhancements}

In this experiment, we augment the base \ourmodel{} architecture with a range of lightweight mechanisms that have shown promise in recent work on efficient sequence modeling. These additions are designed to improve local processing, inductive bias, and compositional mixing.

\paragraph{Setup}
All experiments are conducted in both the \strictsmall{} and \strict{} tracks using the same training setup as in previous sections. The learning rate is fixed at 4e-4, and the Muon optimizer is used for all runs.

Each enhancement is introduced independently to isolate its effect on performance. In addition, a subset of combinations is also evaluated to test potential synergies between modules. Results are reported in terms of validation perplexity and average zero-shot score across the BabyLM benchmark suite.

\begin{table}
    \centering
    \resizebox{0.45\textwidth}{!}{
    \begin{tabular}{lcccc}
    \toprule
    & \multicolumn{2}{c}{\strictsmall} & \multicolumn{2}{c}{\strict} \\
    \textsc{Mechanism} & \textsc{PPL.} & \textsc{Avg.} & \textsc{PPL.} & \textsc{Avg.} \\
    \midrule
    \ourmodel{} & 20.01 & \textbf{37.27} & 7.95 & 35.08 \\
    - \textit{ShortConv} & 12.37 & 36.41 & 6.48 & 34.57 \\
    - \textit{SWA} & 12.08 & 36.16 & 7.38 & 35.86 \\
    - \textit{SWA with Memory} & 10.08 & 34.96 & 6.67 & 37.21 \\
    - \textit{SWA DynMod} & 9.44 & 36.15 & 7.76 & \textbf{38.82} \\
    - \textit{SWA DynMod Bounded} & 8.58 & 34.41 & 6.84 & 36.21 \\
    - \textit{Hedgehog} & 6.18 & 33.58 & 6.68 & 36.65 \\
    - \textit{Hedgehog + SWA} & 7.27 & 36.25 & 6.63 & 34.20 \\
    \bottomrule
    \end{tabular}}
    \caption{Evaluation of lightweight architectural enhancements added to BLaLM. Each mechanism is tested independently on both BabyLM tracks. Results include validation perplexity and average zero-shot performance.}
    \label{tab:mechanism_results}
\end{table}

% \begin{table*}[h]
%     \centering
%     \resizebox{\textwidth}{!}{
%     \begin{tabular}{lcccccccc}
%     \toprule
%     \textsc{Model} & \makecell{\textsc{BLIMP} \\ {\small acc.}} & \makecell{\textsc{B. Suppl.} \\ {\small acc.}} & \makecell{\textsc{Entity} \\ {\small acc.}} & \makecell{\textsc{EWoK} \\ {\small acc.}} & \makecell{\textsc{Eye} \\ {\small $\Delta R^2$}} & \makecell{\textsc{Reading} \\ {\small $\Delta R^2$}} & \makecell{\textsc{WuG} \\ {\small acc.}} & \textsc{Avg.} \\
%     \midrule
%     \textit{10M Track} \\
%     - SWA + DynMod (6)(2e-4) & 61.46 & 58.0 & 25.12 & 51.82 & 1.17 & 0.66 & 70.5 & 38.39 \\
%     % epoch-6 & 67.12 & 54.8 & 29.97 & 53.09 & 1.45 & 0.88 & 66.5 & 39.12 \\
%     - SWA + DynMod (7)(4e-4) & 65.49 & 56.4 & 25.79 & 52.0 & 1.51 & 0.82 & 66.0 & 38.29 \\
% 
%     \midrule
%     \textit{100M Track} \\
%     \ourmodel{} \\
%     - SWA + DynMod (8) & 75.46 & 67.2 & 22.7 & 55.55 & 1.28 & 1.03 & 50.0 & 39.03 \\
%     - SWA + Bounded DynMod (10) & 73.63 & 63.2 & 35.28 & 52.82 & 1.16 & 0.75 & 57.5 & 40.62 \\
%     - SWA + Bounded DynMod + Hedgehog  (8) & 73.62 & 58.8 & 18.3 & 56.82 & 1.11 & 0.48 & 59.5 & 38.38 \\
% 
%     %^ TODO
%     %  /vol/tmp/hallepat/babylm/models/model-2025-08-08_16-42-41/
%     %^ epoch-9 & 75.34 & 62.8 & 23.21 & 54.91 & 0.72 & 0.58 & 55.5 & 39.01 \\
%     \midrule
%     \bottomrule
%     \end{tabular}}
%     \caption{}
%     \label{tab:final_submission}
% \end{table*}

\paragraph{Results}
Table~\ref{tab:mechanism_results} presents the results. In the \strictsmall{} track, most mechanisms improve over the base model, with ShortConv and SWA variants performing particularly well. Hedgehog yields the lowest perplexity (6.18), suggesting improved optimization efficiency, although this does not translate directly into the highest downstream score.

In the \strict{} track, the most effective mechanism is SWA combined with dynamic modulation, which reaches the highest average score of 38.82. Hedgehog and bounded DynMod also improve performance relative to the base configuration.

We additionally tracked the learned weights $\alpha$ for SWA in the dynamic modulation setups. As shown in Appendix~\ref{appendix:mechanisms}, these weights vary across layers and increase over training time, suggesting that deeper layers rely more heavily on local context mixing.

Overall, these results indicate that augmenting mLSTM with lightweight attention or modulation mechanisms can improve both perplexity and downstream performance, particularly when local structure and compositional control are emphasized.

\subsection{Final Submission Models}

For our final BabyLM 2025 submissions, we select configurations that balance strong downstream performance with stable optimization, as identified in our preceding experiments.

\paragraph{\strictsmall{} Track (10M words):} We use BLaLM with mLSTM token mixing, augmented with short convolutions. The learning rate is set to 7e-4, and optimization uses Muon for matrix-shaped parameters and AdamW for scalars. This configuration yields robust zero-shot accuracy across linguistic and educational benchmarks while maintaining low perplexity.

\paragraph{\strict{} Track (100M words):} We adopt the same architecture, but with a learning rate of 5.5e-4, which in our sweep showed superior generalization in higher-data regimes. We include SWA with bounded dynamic modulation, avoiding further additions to preserve architectural simplicity.

We denote the models \ourmodel{}-\strictsmall{} and \ourmodel{}-\strict{} respectively.

In both tracks, models are trained for 10 epochs using the curated dataset described in Section 3. Final submissions are fine-tuned on GLUE for hidden test set evaluation, as per shared task protocol.

The results are shown in Table~\ref{tab:final_model_summary}.

% Strict-Small
% Epoch 10 & 66.45 & 54.8 & 35.96 & 51.27 & 1.07 & 0.97 & -0.21 & 30.04 \ 
% strict
% Epoch 8 & 74.28 & 62.0 & 24.45 & 53.0 & 1.38 & 0.68 & 0.32 & 30.87 \\

\begin{table}[h]
    \centering
    \resizebox{\linewidth}{!}{
    \begin{tabular}{lcc}
    \toprule
    \textsc{Model} & \makecell{\textsc{Zero-Shot} \\ \textsc{Avg.}} & \makecell{\textsc{Fine-Tune} \\ \textsc{Avg.}} \\
    \midrule
    \ourmodel{}-\strictsmall & 29.54 & 57.35 \\
    \ourmodel{}-\strict & 36.49 & 56.70 \\
    \bottomrule
    \end{tabular}}
    \caption{Final performance of our submitted models (\ourmodel{}-\strictsmall{} and \ourmodel{}-\strict{}) on the full BabyLM and (Super)GLUE benchmark suites. Results are averaged across all tasks.}
    \label{tab:final_model_summary}
\end{table}

\section{Conclusion}
We introduced \textbf{\ourmodel{}}, a sample-efficient language model built with linear attention and lightweight enhancements. Across both strict and strict-small tracks, \ourmodel{} outperforms Transformer baselines in low-resource settings and remains competitive at larger scales. Our results highlight two actionable insights: (1) combining mLSTM with sliding window attention and dynamic modulation consistently improves downstream generalization, and (2) the Muon optimizer stabilizes training and reduces perplexity, outperforming AdamW for matrix-valued parameters. These findings offer concrete guidance for efficient model design in data-constrained environments.

\bibliography{anthology,custom}
\bibliographystyle{acl_natbib}

\onecolumn
\appendix

\section{Dataset Curation: Prompt}\label{appendix:prompt}

\begin{figure*}[h]
    \centering
    \begin{minipage}{0.8\textwidth}
    \begin{lstlisting}
Below is an extract from a web page. Evaluate whether the page has a high educational value and could be useful in an educational setting for teaching from primary school to grade school levels using the additive 5-point scoring system described below. Points are accumulated based on the satisfaction of each criterion:
Scoring Criteria:
- +1 Educational Relevance: The extract contains factual or instructional content related to general knowledge, science, math, language, or other academic domains, even if mixed with irrelevant content like ads or unrelated commentary.

- +1 Coherence and Structure: The extract has a recognizable structure (e.g. paragraphs, bullet points, logical flow) and is written in a mostly coherent and syntactically correct way, even if it includes some tangents or inconsistencies.

- +1 Readability and Simplicity: The language is accessible to grade school students, avoiding technical jargon or overly complex sentence constructions. Sentences are clear, concise, and vocabulary is age-appropriate.

- +1 Explainability and Pedagogical Quality: Concepts are explained, not just stated. The text may include analogies, definitions, or examples that make it easier to understand. It supports comprehension and learning.

- +1 Learnability by Small Models: The extract is particularly suitable for training smaller language models: it avoids long-range dependencies, sticks to one or two topics, and has low noise and high signal. Ideal examples follow a pattern, use repetition to reinforce structure, and do not rely heavily on context outside the extract.

The extract:
{0}

After examining the extract: 
- Briefly justify your total score, up to 100 words.
- Conclude with the score using the format: "Educational score:  <total points>"
    \end{lstlisting}
    \end{minipage}%
    \label{fig:enter-label}
    \caption{LLM-based prompt used to assign a \textbf{custom} educational scores to FineWeb-Edu samples. The prompt includes a 5-point additive scoring rubric focusing on pedagogical value, readability, and coherence.}
\end{figure*}

\pagebreak
\section{Model Configurations}\label{appendix:model_configuration}

\begin{table}[h]
    \centering
    \begin{tabular}{lr}
    \toprule
    Hyperparameter & Value \\
    \midrule
    Hidden Size  & 1024 \\
    Intermediate Size & 1536 \\
    Num Attention Heads & 16 \\
    Num Hidden Layers & \\
    - Transformer & 26 \\
    - \ourmodel{} & 24 \\
    Vocab Size & 15K \\
    Parameter Count & \\
    - Transformer & 250M \\
    - \ourmodel{} & 270M \\
    \bottomrule
    \end{tabular}
    \caption{Model configurations for Transformer and BLaLM. Hidden layer count is adjusted to ensure comparable parameter counts across architectures.}
    \label{tab:placeholder}
\end{table}

\section{Experiment 1: Full Results}\label{appendix:dataset_comparison_full_results}

\begin{table*}[ht]
    \centering
    \begin{tabular}{lccccccc}
    \toprule
    \textsc{Dataset} & \makecell{\textsc{BLiMP} \\ {\small acc.}} & \makecell{\textsc{B. Suppl.} \\ {\small acc.}} & \makecell{\textsc{Entity} \\ {\small acc.}} & \makecell{\textsc{EWoK} \\ {\small acc.}} & \makecell{\textsc{Eye} \\ {\small $\Delta R^2$}} & \makecell{\textsc{Reading} \\ {\small $\Delta R^2$}} & \textsc{Avg.} \\
    \midrule
       Baseline-10M (7) & 64.96 & 66.8 & 40.01 & 51.55 & 0.98 & 0.45 & 37.45 \\
       Baseline-10M (8) & 65.49 & 60.4 & 35.22 & 52.27 & 0.76 & 0.4 & 35.75 \\
       Baseline-100M (9) & 75.43 & 63.6 & 20.47 & 53.36 & 0.59 & 0.26 & 35.61 \\
       Baseline-100M (6) & 75.68 & 65.2 & 34.82 & 51.82 & 0.82 & 0.33 & 38.11 \\
    \midrule
       Our-10M (8) & 67.99 & 63.6 & 39.46 & 52.27 & 1.09 & 0.65 & 37.51 \\
       Our-10M (9) & 66.81 & 59.2 & 41.97 & 52.73 & 0.85 & 0.53 & 37.01 \\
       Our-100M (10) & 74.51 & 57.6 & 14.65 & 55.73 & 1.18 & 0.9 & 34.09 \\
       Ours-100M (10) & 74.6 & 57.2 & 16.56 & 53.64 & 1.21 & 0.57 & 33.96 \\
    \bottomrule
    \end{tabular}
    \caption{Detailed results comparing our curated dataset to the official BabyLM baseline. The number in parentheses indicates the best-performing training epoch.}
    \label{tab:dataset_comparison_full_results}
\end{table*}

\pagebreak
\section{Experiment 2: Full Results}\label{appendix:architecture_comparison_full_results}

\begin{table*}[ht]
    \centering
    \resizebox{\textwidth}{!}{
    \begin{tabular}{cll|ccccccc}
    \toprule
    \textsc{Track} & \textsc{Model} & \textsc{LR} & \textsc{BLiMP} & \textsc{B. Suppl.} & \textsc{Entity} & \textsc{EWoK} & \textsc{Eye} & \textsc{Reading} & \textsc{Avg.} \\
     & & & acc. & acc. & acc. & acc. & $\Delta R^2$ & $\Delta R^2$ & \\
    \midrule
    \multirow{6}{*}{Strict-Small}
        & Transformer (10) & 3e-4 & 64.47 & 58.8 & 15.71 & 49.45 & 1.06 & 0.56 & 31.67 \\
        & Transformer (9) & 4e-4 & 64.95 & 57.2 & 18.07 & 51.36 & 1.13 & 0.93 & 32.27 \\
        & Transformer (10) & 5e-4 & 65.37 & 54.4 & 14.5  & 52.18 & 0.83 & 0.44 & 31.28 \\
        & \ourmodel{} (9)    & 3e-4 & 63.07 & 59.6 & 38.52 & 51.27 & 0.82 & 0.5  & 35.63 \\
        & \ourmodel{} (9)    & 4e-4 & 66.93 & 55.6 & 28.74 & 52.09 & 0.91 & 0.5  & 34.12 \\
        & \ourmodel{} (9)    & 5e-4 & 66.72 & 55.6 & 40.93 & 51.0  & 0.91 & 0.6  & 35.96 \\
    \midrule
    \multirow{6}{*}{Strict}
        & Transformer (10) & 3e-4 & 72.81 & 60.8 & 18.51 & 55.09 & 0.82 & 0.58 & 34.76 \\
        & Transformer (9)  & 4e-4 & 72.44 & 62.0 & 20.63 & 53.36 & 1.02 & 0.74 & 35.03 \\
        & Transformer (10) & 5e-4 & 72.47 & 63.2 & 18.21 & 51.55 & 1.11 & 0.71 & 34.54 \\
        & \ourmodel{} (10)   & 3e-4 & 73.40 & 61.2 & 14.70 & 54.55 & 0.93 & 0.51 & 34.21 \\
        & \ourmodel{} (10)   & 4e-4 & 74.60 & 57.2 & 16.56 & 53.64 & 1.21 & 0.57 & 33.96 \\
        & \ourmodel{} (10)   & 5e-4 & 74.49 & 60.4 & 21.99 & 53.91 & 1.03 & 0.71 & 35.42 \\
    \bottomrule
    \end{tabular}
    }
    \caption{Detailed zero-shot results for Transformer and BLaLM across BabyLM benchmarks. Parentheses indicate best-performing epoch.}
    \label{tab:architecture_comparison_full_results}
\end{table*}

\section{Experiment 3: Full Results}\label{appendix:optimizer_results}

\begin{table*}[h]
    \centering
    \resizebox{\textwidth}{!}{
    \begin{tabular}{lcccccccc}
    \toprule
    \textsc{Model} & \makecell{\textsc{BLiMP} \\ {\small acc.}} & \makecell{\textsc{B. Suppl.} \\ {\small acc.}} & \makecell{\textsc{Entity} \\ {\small acc.}} & \makecell{\textsc{EWoK} \\ {\small acc.}} & \makecell{\textsc{Eye} \\ {\small $\Delta R^2$}} & \makecell{\textsc{Reading} \\ {\small $\Delta R^2$}} & \textsc{Avg.} \\
    \midrule
    \textit{AdamW} \\
    Run1 (10) & 74.96 & 60.8 & 13.25 & 55.55 & 0.85 & 0.61 & 34.33 \\
    Run2 (10) & 74.37 & 65.6 & 32.8 & 54.82 & 0.9 & 0.68   & 38.19 \\
    Run3 (10) & 76.54 & 61.6 & 14.72 & 53.82 & 0.92 & 0.57 & 34.69 \\

    \midrule
    \textit{Muon} \\
    Run1 (10) & 76.4 & 70.4 & 22.27 & 55.91 & 1.17 & 0.79  & 37.82 \\
    Run2 (10) &  75.82 & 66.4 & 15.39 & 55.73 & 1.1 & 0.88 & 35.88 \\
    Run3 (10) & 75.27 & 64.0 & 15.98 & 53.18 & 1.07 & 0.76 & 35.03 \\
    \bottomrule
    \end{tabular}}
    \caption{Zero-shot results for xLSTM models trained with AdamW and Muon optimizers (3 runs). Parentheses indicate best-performing epoch.}
\end{table*}

\newpage
\section{Experiment 4: Full Results}\label{appendix:lr_sweep}

\begin{table*}[h]
    \centering
    \resizebox{\textwidth}{!}{
    \begin{tabular}{lccccccc}
    \toprule
    \textsc{Learning Rate} & \makecell{\textsc{BLiMP} \\ {\small acc.}} & \makecell{\textsc{B. Suppl.} \\ {\small acc.}} & \makecell{\textsc{Entity} \\ {\small acc.}} & \makecell{\textsc{EWoK} \\ {\small acc.}} & \makecell{\textsc{Eye} \\ {\small $\Delta R^2$}} & \makecell{\textsc{Reading} \\ {\small $\Delta R^2$}} & \textsc{Avg.} \\
    \midrule
    \textbf{\strictsmall} \\
    1e-4 (6) & 57.16 & 58.4 & 39.95 & 52.73 & 0.28 & 0.3  & 36.04 \\
    2e-4 (6) & 61.37 & 59.2 & 41.81 & 52.55 & 0.74 & 0.59 & 34.80 \\
    3e-4 (7) & 67.19 & 60.8 & 33.44 & 50.55 & 1.14 & 0.58 & 35.61 \\
    4e-4 (8) & 69.55 & 58.0 & 41.91 & 52.36 & 1.17 & 0.68 & 37.27 \\
    5e-4 (6) & 69.12 & 59.6 & 26.8  & 52.91 & 1.1  & 0.69 & 35.03 \\
    6e-4 (9) & 69.98 & 61.6 & 18.89 & 52.91 & 1.03 & 0.65 & 34.17 \\
    7e-4 (7) & 70.68 & 60.4 & 38.87 & 53.82 & 0.96 & 0.47 & 37.53 \\
    \midrule
    \textbf{\strict} \\
    2e-4 (9) & 73.81 & 60.4 & 17.42 & 52.64 & 0.87 & 0.54  & 34.28 \\
    3e-4 (10) & 75.91 & 67.6 & 15.06 & 55.09 & 0.81 & 0.5  & 35.82 \\
    4e-4 (8) & 66.82 & 54.8 & 33.54 & 54.0 & 0.92 & 0.43  & 35.08 \\
    5e-4 (10) & 76.25 & 66.4 & 15.08 & 55.55 & 1.03 & 0.62 & 35.82 \\
    5.5e-4 (9) & 76.1 & 63.6 & 28.04 & 55.64 & 0.87 & 0.73 & 37.49 \\
    6e-4 (9) & 76.42 & 59.2 & 18.64 & 54.36 & 0.94 & 0.83  & 35.06 \\
    7e-4 (9) & 75.85 & 66.0 & 19.46 & 53.73 & 1.03 & 0.55  & 36.10 \\
    \bottomrule
    \end{tabular}}
    \caption{Full results for Experiment 4. The number in brackets after each learning rate denotes the best performing epoch.}
\end{table*}

\newpage
\section{Experiment 5: Full Results}\label{appendix:mechanisms}

\begin{table*}[h]
    \centering
      \resizebox{\textwidth}{!}{
    \begin{tabular}{lcccccccc}
    \toprule
    \textsc{Mechanism} & \makecell{\textsc{BLiMP} \\ {\small acc.}} & \makecell{\textsc{B. Suppl.} \\ {\small acc.}} & \makecell{\textsc{Entity} \\ {\small acc.}} & \makecell{\textsc{EWoK} \\ {\small acc.}} & \makecell{\textsc{Eye} \\ {\small $\Delta R^2$}} & \makecell{\textsc{Reading} \\ {\small $\Delta R^2$}} & \textsc{Avg.} \\
    \midrule
    \textbf{\strictsmall} \\
    ShortConv (6)          & 67.13 & 57.6 & 38.82 & 52.82 & 1.42 & 0.71 & 36.41 \\
    SWA (9)                & 64.86 & 54.4 & 43.17 & 52.91 & 1.1  & 0.52 & 36.16 \\
    SWA With Memory (7)    & 65.83 & 52.8 & 35.97 & 52.18 & 1.99 & 1.04 & 34.96 \\
    SWA DynMod (6)         & 67.36 & 54.4 & 41.28 & 51.36 & 1.69 & 0.83 & 36.15 \\
    SWA DynMod Bounded (7) & 65.59 & 52.4 & 33.86 & 52.36 & 1.35 & 0.93 & 34.41 \\
    Hedgehog (10)          & 68.3  & 53.2 & 24.5  & 53.0  & 1.52 & 0.99 & 33.58 \\ 
    Hedgehog + SWA (6)     & 65.43 & 54.4 & 42.49 & 52.73 & 1.55 & 0.94 & 36.25 \\
    \midrule
    \textbf{\strict} \\
    ShortConv (8)           & 74.31 & 61.2 & 15.63 & 54.18 & 1.05 & 1.08 & 34.57 \\
    SWA (8)                 & 74.29 & 60.8 & 21.42 & 56.45 & 1.2 & 1.02  & 35.86 \\
    SWA With Memory (10)    & 71.52 & 65.2 & 31.04 & 54.18 & 0.85 & 0.49 & 37.21 \\
    SWA DynMod (9)          & 76.39 & 68.0 & 31.32 & 55.64 & 0.83 & 0.76 & 38.82 \\
    SWA DynMod Bounded (10) & 73.64 & 66.0 & 22.36 & 53.55 & 0.97 & 0.75 & 36.21 \\
    Hedgehog (8)            & 74.64 & 62.0 & 24.69 & 56.73 & 1.32 & 0.55 & 36.65 \\
    Hedgehog + SWA (7)      & 72.94 & 58.8 & 16.28 & 54.64 & 1.69 & 0.9  & 34.20 \\
    \bottomrule
    \end{tabular}}
    \caption{Full results for Experiment 5. Parentheses indicate the best-performing epoch per configuration.}
\end{table*}

\pagebreak
\subsection{Alpha Value Development for DynMod Runs}

\begin{figure*}[h]
    \includegraphics[width=\textwidth]{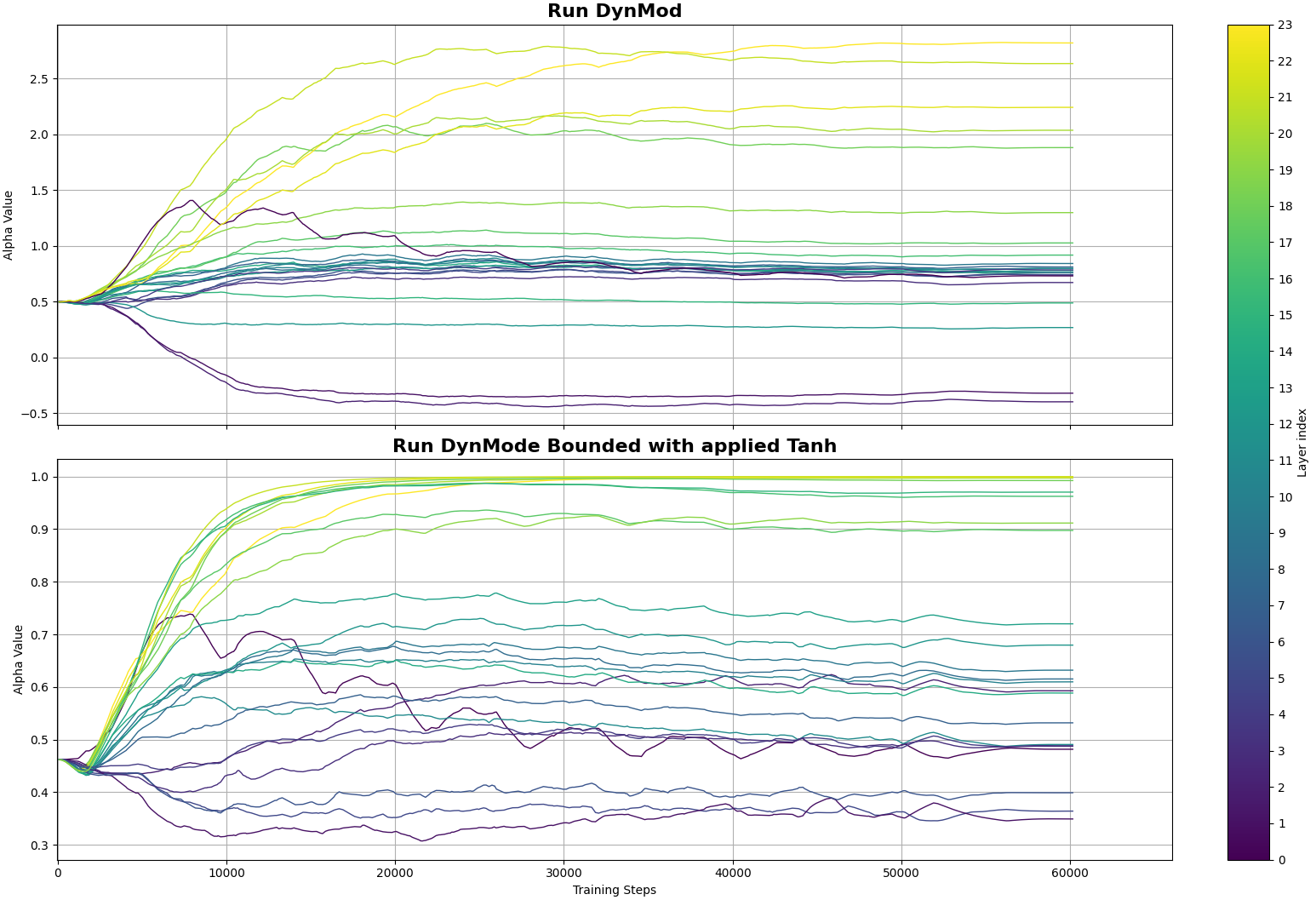}
    \caption{Layer-wise development of dynamic modulation weights ($\alpha$) during training for the bounded DynMod variant. We apply the tanh function to stabilize values. Later layers show increased reliance on local mixing.}
    \label{fig:alpha_values}
\end{figure*}

\pagebreak
\section{Final Submission: Full Results}
\begin{table*}[h]
    \centering
    \begin{tabular}{lcc}
    \toprule
    \textsc{Benchmark} & \makecell{\ourmodel{} \\ \strictsmall} & \makecell{\ourmodel{} \\ \strict }\\
    \midrule
    \textsc{BLiMP} & 67.0 & 74.7 \\
    \textsc{B. Suppl.} & 53.3 & 61.0 \\
    \textsc{Entity Tracking} & 33.7 & 22.2 \\
    \textsc{EWoK} & 50.6 & 53.6 \\
    \textsc{Eye Tracking} & 1.1 & 1.1 \\
    \textsc{Self Paced Reading} & 1.0 & 0.6 \\
    \textsc{WUG Adj. Norm.} & 50.3 & 47.5 \\
    \textsc{WUG. Past Tense} & -20.7 & 37.5 \\
    \textsc{COMPS} & 50.5 & 58.3 \\
    \textsc{AoA} & 8.6 & 8.6 \\
    \midrule
    \textsc{Average} & 29.54 & 36.49 \\
    \bottomrule
    \end{tabular}
    \caption{Final BabyLM benchmark results for \ourmodel{}-\strictsmall and \ourmodel{}-\strict models. Includes hidden tasks.}
    \label{tab:final_models_zero_shot}
\end{table*}

\begin{table*}[h]
    \centering
    \resizebox{\textwidth}{!}{
    \begin{tabular}{lcccccccc}
    \toprule
    \textsc{Model} & \makecell{\textsc{BoolQ} \\ {\small acc.}} & \makecell{\textsc{MNLI} \\ {\small acc.}} & \makecell{\textsc{MRPC} \\ {\small acc.}} & \makecell{\textsc{QQP} \\ {\small acc.}} & \makecell{\textsc{MultiRC} \\ {\small acc.}} & \makecell{\textsc{RTE} \\ {\small acc.}} & \makecell{\textsc{WSC} \\ {\small acc.}} & \textsc{Avg.} \\
    \midrule
    \ourmodel{}-\strictsmall & 64.03 & 34.18 & 69.60 & 59.92 & 57.54 & 54.67 & 61.54 & 57.35 \\
    \ourmodel{}-\strict & 64.03 & 34.27 & 69.10 & 58.90 & 57.54 & 51.79 & 61.53 & 56.70 \\
    \bottomrule
    \end{tabular}}
    \caption{Performance of \ourmodel{}-\strictsmall and \ourmodel{}-\strict on (Super)GLUE tasks after fine-tuning.}
    \label{tab:final_models_glue}
\end{table*}

\end{document}